\documentclass[letterpaper, 10 pt, conference]{ieeeconf}  
\IEEEoverridecommandlockouts                              
\usepackage{amsmath}
\usepackage{amssymb}
\usepackage{gensymb}
\usepackage{algorithm} 
\usepackage{algpseudocode}
\usepackage{url}
\usepackage{graphicx}
\usepackage{float}
\usepackage{setspace}
\usepackage{subfigure}
\usepackage{xcolor}
\usepackage{wrapfig}
\DeclareMathOperator*{\argmax}{argmax}
\DeclareMathOperator{\sign}{sign}

\renewcommand{\baselinestretch}{0.965}

\newtheorem{problem}{Problem}

\title{\LARGE \bf
Classification of Time-Series Data Using Boosted Decision Trees
}

\author{Erfan Aasi$^{1}$, Cristian Ioan Vasile$^{2}$, Mahroo Bahreinian$^{1}$, and Calin Belta$^{1}$
\thanks{*This work was partially supported by the NSF under grants IIS-2024606 and IIS-1723995 at Boston University.}
\thanks{$^{1}$Erfan Aasi, Calin Belta, and Mahroo Bahreinian are with
        Boston University, 
        Boston, MA 02215, USA
        {\tt\small eaasi@bu.edu}, {\tt\small cbelta@bu.edu}, {\tt\small mahroobh@bu.edu}}
\thanks{$^{2}$Cristian Ioan Vasile is with Lehigh University,
        Bethlehem, PA 18015, USA
        {\tt\small cvasile@lehigh.edu}}
}

\begin{document}

\maketitle
\thispagestyle{empty}
\pagestyle{empty}

\begin{abstract}
Time-series data classification is central to the analysis and control of autonomous systems, such as robots and self-driving cars. Temporal logic-based learning algorithms have been proposed recently as classifiers of such data. However, current frameworks are either inaccurate for real-world applications, such as autonomous driving, or they generate long and complicated formulae that lack interpretability. To address these limitations,  we introduce a novel learning method, called Boosted Concise Decision Trees (BCDTs), to generate binary classifiers that are represented as Signal Temporal Logic (STL) formulae. Our algorithm leverages an ensemble of Concise Decision Trees (CDTs) to improve the classification performance, where each CDT is a decision tree that is empowered by a set of techniques to generate simpler formulae and improve interpretability.
The effectiveness and classification performance of our algorithm are evaluated on naval surveillance and urban-driving case studies.
\end{abstract}

\section{INTRODUCTION} \label{sec:introduction}
To cope with the complexity of robotic tasks,
machine learning (ML) techniques have been employed to capture their
temporal and logical structure from time-series data. One of the main problems in ML is the two-class classification problem, where the goal is to build a classifier that distinguishes desired system behaviors from the undesired ones. Traditional ML algorithms focus on building such classifiers; however, they are often not easy to understand or they don't offer any insights about the system. Motivated by the readability and interpretability of temporal logic formulae~\cite{clarke1986automatic}, there has been great interest in applying formal methods to ML in recent years \cite{asarin2011parametric,jha2019telex,vazquez2017logical,neider2018learning,xu2019information,ketenci2019synthesis, yan2021neural}.

Signal Temporal Logic (STL) \cite{maler2004monitoring} is a specification language used to express temporal properties of real-valued signals. In this paper, we use STL to generate specifications of time-series system behaviors. Early methods for mining temporal properties from data mostly focus on parameter synthesis, given template formulae~\cite{asarin2011parametric,jin2015mining,hoxha2018mining,bakhirkin2018efficient}. These works require the designer to have a good understanding of the system properties. 
In~\cite{kong2016temporal}, a general supervised learning framework that infers both the structure and the parameters of a formula is presented. The approach is based on lattice search and parameter synthesis,
which makes it general, but inefficient.
Using an efficient decision tree-based framework to learn STL formulae is explored in~\cite{bombara2016decision, bombara2021offline}, where the nodes of the tree contain simple formulae that are tuned optimally from a predefined set of primitives. In \cite{mohammadinejad2020interpretable}, the authors propose a systematic enumeration based method to learn short, interpretable STL formulae.
Other works about learning temporal logic formulae consider learning from positive examples only~\cite{jha2019telex}, clustering~\cite{vazquez2017logical},
active learning~\cite{linard2020active}, and automata-based methods for untimed formulae~\cite{neider2018learning,xu2019information}. 

 Most existing algorithms for learning STL formulae either do not achieve good classification performance for real-world applications, or do not provide any interpretability of the output formulae: they generate long and complicated specifications. To address these concerns, in this paper we introduce \emph{Boosted Concise Decision Trees} (BCDTs) to learn STL formulae from labeled time-series data. To improve the classification accuracy of existing works, we use a boosting method to combine multiple models with weak classification power. The weak learning models are bounded-depth decision trees, called Concise Decision Trees (CDTs). Each CDT is a Decision Tree (DT) \cite{breiman1984classification}, empowered by a set of techniques called conciseness techniques, to generate simpler formulae and improve the interpretability of the final output. We also use a heuristic method in the BCDT algorithm to prune the ensemble of trees, which helps with the interpretability of the formulae. To relate STL and BCDTs, we establish a connection between boosted trees and weighted STL (wSTL) formulae \cite{mehdipour2020specifying}, which have weights associated with Boolean and temporal operators. We show performance gains and improved interpretability of our method compared to literature, in naval surveillance and urban driving scenarios. 

The main contributions of the paper are: (a) a novel inference algorithm based on boosted decision trees, which has better classification performance than related approaches, (b) a set of heuristic techniques to generate simple STL formulae from decision trees that improve interpretability, (c) two case studies in naval surveillance and urban-driving that highlight the classification performance and interpretability of our proposed learning algorithm.

\section{PRELIMINARIES} \label{sec:preliminaries}
Let $\mathbb{R}$, $\mathbb{Z}$, $\mathbb{Z}_{\geq 0}$ denote the sets of real, integer, and non-negative integer numbers, respectively.
With a slight abuse of notation, given 
$a,b\in \mathbb Z_{\geq0}$
we use $[a,b]=\{t\in\mathbb Z_{\geq0}\ |\ a\leq t\leq b\}$.
The cardinality of a set is denoted by $|\cdot|$.
A (discrete-time) signal $s$ is a function $s : [0, T] \to \mathbb{R}^n$
that maps each (discrete) time point $t \in [0, T]$
to an $n$-dimensional vector of real values, where $T \in \mathbb{Z}_{\geq 0}$.
Each component of $s$ is denoted as $s_j, j \in [1, n]$.

Signal Temporal Logic (STL) was introduced in~\cite{maler2004monitoring}. 
Informally, the STL formulae used in this paper are made of predicates $\mu$ defined over components of real-valued signals in the form of
$\mu = s_j \sim \pi$, where $\pi \in \mathbb{R}$ is a threshold and $\sim \in \{>, \leq \}$, which are connected using Boolean operators, such as $\lnot$, $\land$, $\lor$, and temporal operators, such as $G_{[a,b]}$ (always) and $F_{[a,b]}$ (eventually).
The semantics are defined over signals. For example, formula $G_{[3,6]}s_3 \leq1$ means that, for all times 3,4,5,6,
component $s_3$ of a signal $s$ is less than or equal 1.
STL has both qualitative and quantitative semantics.
We use $s\models\phi$ to denote Boolean satisfaction.
The quantitative semantics is given by a robustness degree $\rho(\phi, s)$\cite{donze2010robust} , which captures the degree of satisfaction 
of a formula $\phi$ by a signal $s$.
Positive robustness ($\rho(\phi, s) \geq 0$) implies Boolean satisfaction $s\models\phi$, while negative robustness ($\rho(\phi, s) < 0$) implies violation $s\not\models \phi$.

Weighted STL (wSTL) \cite{mehdipour2020specifying} is an extension of STL that has the same qualitative semantics as STL, but has weights associated with the Boolean and temporal operators, which modulate its robustness degree. In this paper, we restrict our attention to a fragment of wSTL with weights on conjunctions only. For example, the wSTL formula $\phi_1 \land^{\alpha} \phi_2$, $\alpha=(\alpha_1, \alpha_2) \in \mathbb{R}^2_{>0}$,
denotes that $\phi_1$ and $\phi_2$ must hold with priorities $\alpha_1$ and $\alpha_2$. The priorities capture the satisfaction importance of their corresponding formulae.

Parametric STL (PSTL) \cite{asarin2011parametric} is an extension of STL, where the endpoints $a,b$ of the time intervals in the temporal operators and the thresholds $\pi$ in the predicates are parameters.
The set of all possible valuations of all parameters in a PSTL formula $\psi$ is called the parameter space and is denoted by 
$\Theta$. A particular valuation is denoted by $\theta\in\Theta$ and the corresponding formula by $\psi_\theta$.

\section{PROBLEM FORMULATION} \label{sec:prob-form}

\subsection{Motivating Example} \label{sec:motivating-example}
\begin{figure} [htb]
    \centering
    \subfigure[]
    {\includegraphics[width=0.45\columnwidth, height=0.35\columnwidth]{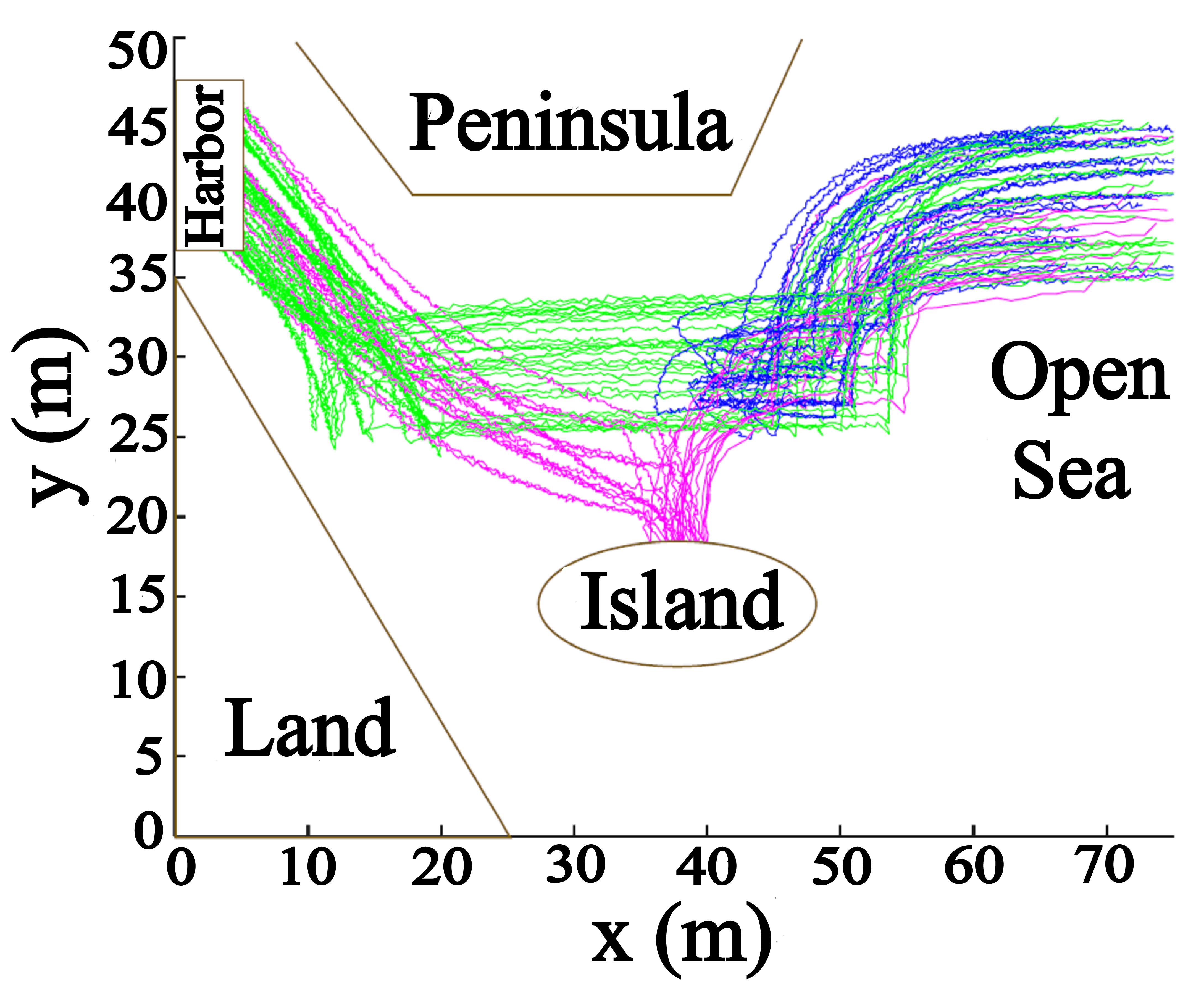}\label{fig:naval-scenario}}
    \subfigure[]
    {\includegraphics[width=0.45\columnwidth, height=0.35\columnwidth]{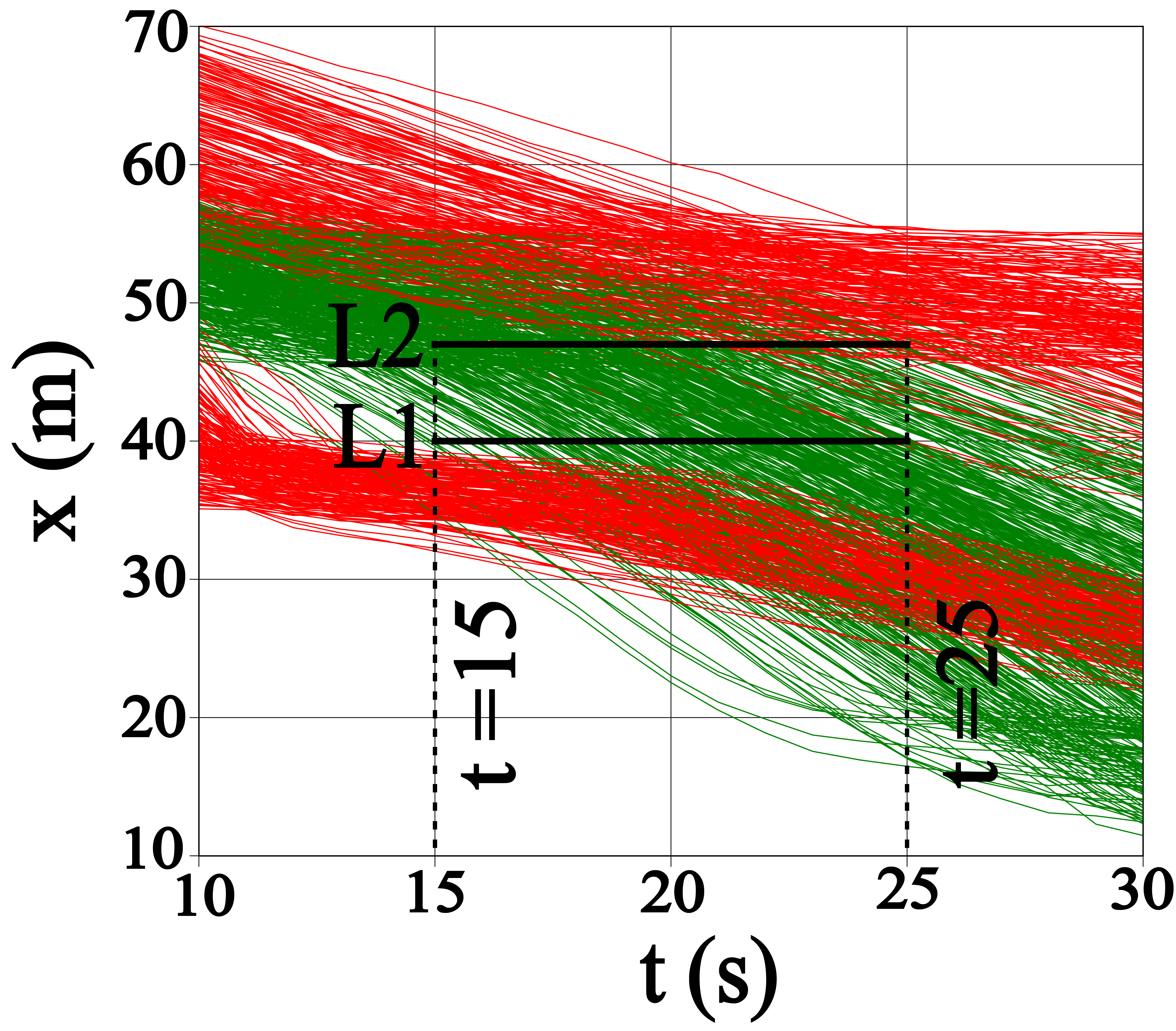}\label{fig:naval-x}}
    \subfigure[]
    {\includegraphics[width=0.45\columnwidth, height=0.35\columnwidth]{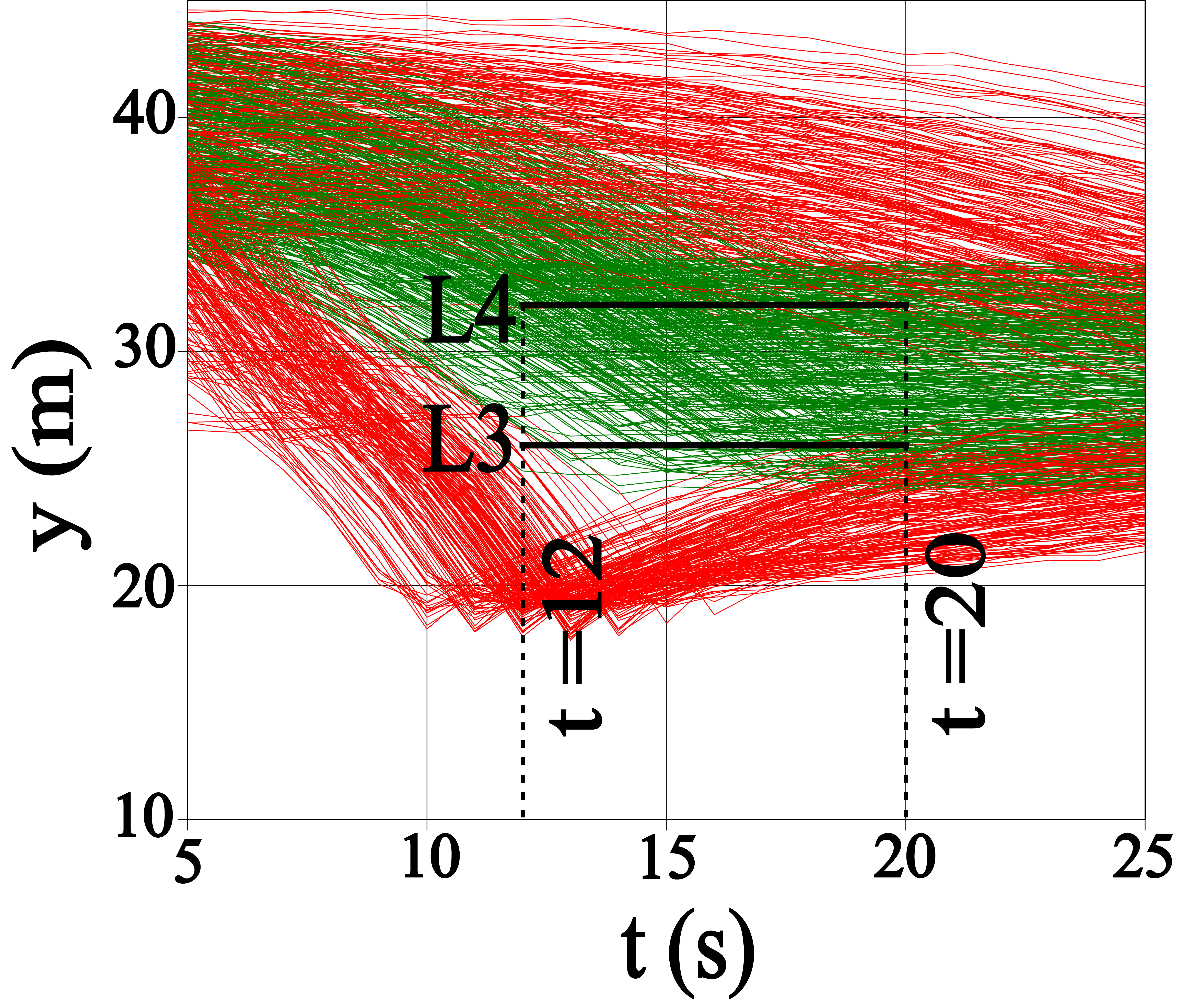}\label{fig:naval-y}} 
    \caption{(a) Naval surveillance scenario \cite{kong2016temporal}, where normal trajectories are shown in green, and anomalous signals are shown in blue and magenta, (b) x and (c) y components of naval trajectories. The green and red trajectories belong to the normal and anomalous behaviors, respectively.}
\vspace{-2mm}
\end{figure}

Consider the maritime surveillance scenario from \cite{kong2016temporal,bombara2021offline} (see Fig.~\ref{fig:naval-scenario}). The goal is to detect anomalous vessel behaviors by looking at their trajectories. A vessel behaving normally approaches from the open sea and heads directly towards the harbor, while a vessel with anomalous behaviors either veers to the island and then heads to the harbor, or it approaches other vessels in the passage between the peninsula and the island and then returns to the open sea. 


In the scenario's dataset \cite{bombara2021offline}, the signals are represented as 2-dimensional trajectories with planar coordinates $(x(t), y(t))$.
The labels indicate the type of a vessel's behavior (normal or anomalous).
In Fig.~\ref{fig:naval-x} and~\ref{fig:naval-y}, we show the $x$ and $y$ components of some signals, respectively, over time. For better visualization, we show the signals over a part of their time horizon. In Fig.~\ref{fig:naval-x}, one of the areas that distinguishes between positive (normal) and negative (anomalous) signals is the area between lines $L1$ and $L2$, over the time interval $t \in [15, 25]$. By using STL classifiers, formula $\phi_1 = F_{[15, 25]} (x > 40) \wedge F_{[15, 25]} (x \leq 47)$, or even a simpler formula $\phi'_1 = F_{[15, 25]} ((x > 40) \wedge (x \leq 47))$, can be used to distinguish positive and negative signals in this area. Similarly, in Fig.~\ref{fig:naval-y}, we can describe the separation area between lines $L3$ and $L4$ by the STL formula $\phi_2 = F_{[12, 20]} (y > 26) \wedge F_{[12, 20]} (y \leq 32)$, or even a simpler formula $\phi'_2 = F_{[12, 20]} ((y > 26) \wedge (y \leq 32))$. Considering the common time interval between the separation areas in Fig.~\ref{fig:naval-x} and Fig.~\ref{fig:naval-y}, a shorter and easier to read formula $\phi_3 = F_{[15, 20]} ((40 < x \leq 47) \wedge (26 < y \leq 32))$ can be used to distinguish between positive and negative signals in the x-y space.
In this paper, we seek techniques to generate simple formulae, such as $\phi_3$, to classify signals without losing the classification accuracy.

\subsection{Problem Statement}
Let $C = \{C_p, C_n\}$ be the set of possible (positive and negative) classes. We consider a labeled data set with $N$ data samples as $S = \{(s^i, \ell^i) \}_{i=1}^{N}$, where $s^i$ is the $i^{\text{th}}$ signal and $\ell^i \in C$ is its label.
\vspace{2mm}
\begin{problem}\label{probform:formula-learning}
Given a labeled data set $S= \{(s^i, \ell^i)\}_{i = 1}^{N}$,
find an STL formula $\phi$ that minimizes the Misclassification Rate $MCR(\phi)$ defined below:
\begin{equation} \label{eqn:non-incproblem}
    \frac{|\{s^i \mid (s^i \models \phi \, \wedge \, \ell^i = C_n ) \, \vee \, (s^i \nvDash \phi \, \wedge \, \ell^i = C_p) \}|}{N}
\end{equation}
\end{problem}
\vspace{2mm}

\section{SOLUTION} \label{section:solution}
We propose a solution to Pb.~\ref{probform:formula-learning} based on BCDT method, presented in Sec.~\ref{sec:boosted_trees}. BCDT grows multiple binary CDTs, inspired by AdaBoost~\cite{freund1997decision} algorithm, where each CDT is a decision tree empowered by a set of conciseness techniques to generate simpler formulae. The construction method for a single CDT is explained in Sec.~\ref{sec:single_trees}. We describe the meta parameters of the CDT method in Sec.~\ref{sec:meta_params}, and in  Sec.~\ref{sec:expressivity_methods} we explain the conciseness techniques and the connection with interpretability.

\subsection{Boosted Concise Decision Trees Algorithm} \label{sec:boosted_trees}
The BCDT algorithm in Alg.~\ref{alg:boostedtree} is inspired by the AdaBoost method \cite{shalev2014understanding}. AdaBoost combines weak classifiers with simple formulae, trained on weighted data samples. Weights of the data represent the difficulty of correct classification. After training a weak classifier, the weights of the correctly classified samples are decreased and weights of the misclassified samples are increased. The algorithm takes as input the labeled data set $S$,  the number of learners (trees) $K$, and the weak learning model $\mathcal{E}$, which is the algorithm to construct CDTs (explained in Alg.~\ref{alg:dec_tree}). The CDTs are binary decision trees, where formulae of the nodes are primitives (see Sec.~\ref{sec:meta_params}) with general rectangular predicates $\mu$ of the form $A s \leq b$,
with $A = [\mathbb{I}_{n_1}\ \, -\mathbb{I}_{n_2}]^T$, $b \in \mathbb{R}^{n_1 + n_2}$, $\mathbb{I}_n$ as the $n \times n$ identity matrix, and $n_1, n_2 \in [0, n]$.

\begin{algorithm}[htb]
\caption{Boosted Concise Decision Trees (BCDT)}
\begin{algorithmic}[1]
    \State \textbf{Input:} $S = \{(s^i, \ell^i)\}_{i = 1}^{N}$, $K$, $\mathcal{E}$ 
    \State \textbf{Output:} final classifier $f_{BCDT}(\cdot)$
    \State \textbf{Initialize:} $\forall \, (s^i,\ell^i) \in S: \, D_1(s^i) = 1/|S|$ \label{lst:line:initialize}
    \State \textbf{for} k = 1, $\ldots$, K: \label{lst:line:loop}
    \State \hskip1.5em $\text{classifier} \, f_{CDT}^{k}(\cdot) \leftarrow \mathcal{E}(S, D_k)$  \label{lst:line:sinlgtree}
    \vspace{1mm}
    \State \hskip1.5em $\epsilon_k \gets \sum_{(s^i, \ell^i) \in S} D_k(s^i) \, \cdot \, \mathbf{1}[\ell^i \neq f_{CDT}^{k}(s^i)]$ \label{lst:line:misclasserror}
    \vspace{1mm}
    \State \hskip1.5em $\alpha_k \gets  \begin{cases} \frac{1}{2} \ln{(\frac{1}{\epsilon_k} - 1)} &  0 < \epsilon_k \leq 1/2\\
    M & \epsilon_k = 0
    \end{cases}$ \label{lst:line:treeweight}
    \vspace{1mm}
    \State \hskip1.5em $D_{k+1}(s^i) \propto D_k(s^i) \exp{(-\alpha_k \cdot \ell^i \cdot f_{CDT}^{k}(s^i))}$ \label{lst:line:weightupdate}
    \vspace{1mm}
    \State {\small $f_{BCDT}(\cdot) \gets \begin{cases} \sign(\sum_{k=1}^{K} \alpha_k \cdot f_{CDT}^{k}(\cdot)) & \alpha_k < M, \forall k\\
    f_{CDT}^{k^*}(\cdot) & \text{otherwise}
    \end{cases}$} \label{lst:line:output}
    \vspace{1mm}
    \State \textbf{return} $f_{BCDT}(\cdot)$
\end{algorithmic}
\label{alg:boostedtree}
\end{algorithm} 

In Alg.~\ref{alg:boostedtree}, initially all data samples are weighted equally (line~\ref{lst:line:initialize}).
The algorithm iterates over the number of trees (line~\ref{lst:line:loop}).
At each iteration, the weak learning algorithm $\mathcal{E}$
constructs a single CDT $f_{CDT}^{k}(\cdot)$
based on data set $S$ and current samples' weights $D_k$
(line~\ref{lst:line:sinlgtree}).
Next, the misclassification error of the constructed tree $\epsilon_k$ is computed (line~\ref{lst:line:misclasserror}). If the current tree has weak classification performance better than random guessing ($0 < \epsilon_k \leq 1/2$), its weight is computed based on the original AdaBoost method, and if it has perfect classification performance and classifies all signals correctly ($\epsilon_k = 0$), a big value $M$ is assigned to its weight (line~\ref{lst:line:treeweight}). 
At the end of each iteration, the samples' weights are updated and normalized (denoted by $\propto$)
based on the performance of the current tree, to focus on the misclassified signals in the next trees (line~\ref{lst:line:weightupdate}). 
To compute the final output of the algorithm, we use a heuristic method to prune the ensemble of trees, to generate simpler formulae and improve interpretability. Inspired by heuristic methods for pruning ensemble of decision trees in \cite{breiman1984classification, kulkarni2012pruning}, we compute the final output $f_{BCDT}(\cdot)$ as (line~\ref{lst:line:output}): if the weights of all trees are less than $M$, the final output is computed as the weighted majority vote over all the CDTs (as in the AdaBoost method); otherwise, if there are one or more trees with weight $M$, the final output is computed by the tree with weight $M$ that has the simplest STL formula, denoted by $f_{CDT}^{k^*}(\cdot)$. As a metric to compare the simplicity of formulae, the number of Boolean and temporal operators is considered. This pruning method helps with reducing the generalization error in the test phase and generating simpler formulae. We show its advantages with empirical results in Sec.~\ref{section:case-studies}.

\begin{figure}[h]
\centering
\includegraphics[width=1.0\columnwidth]{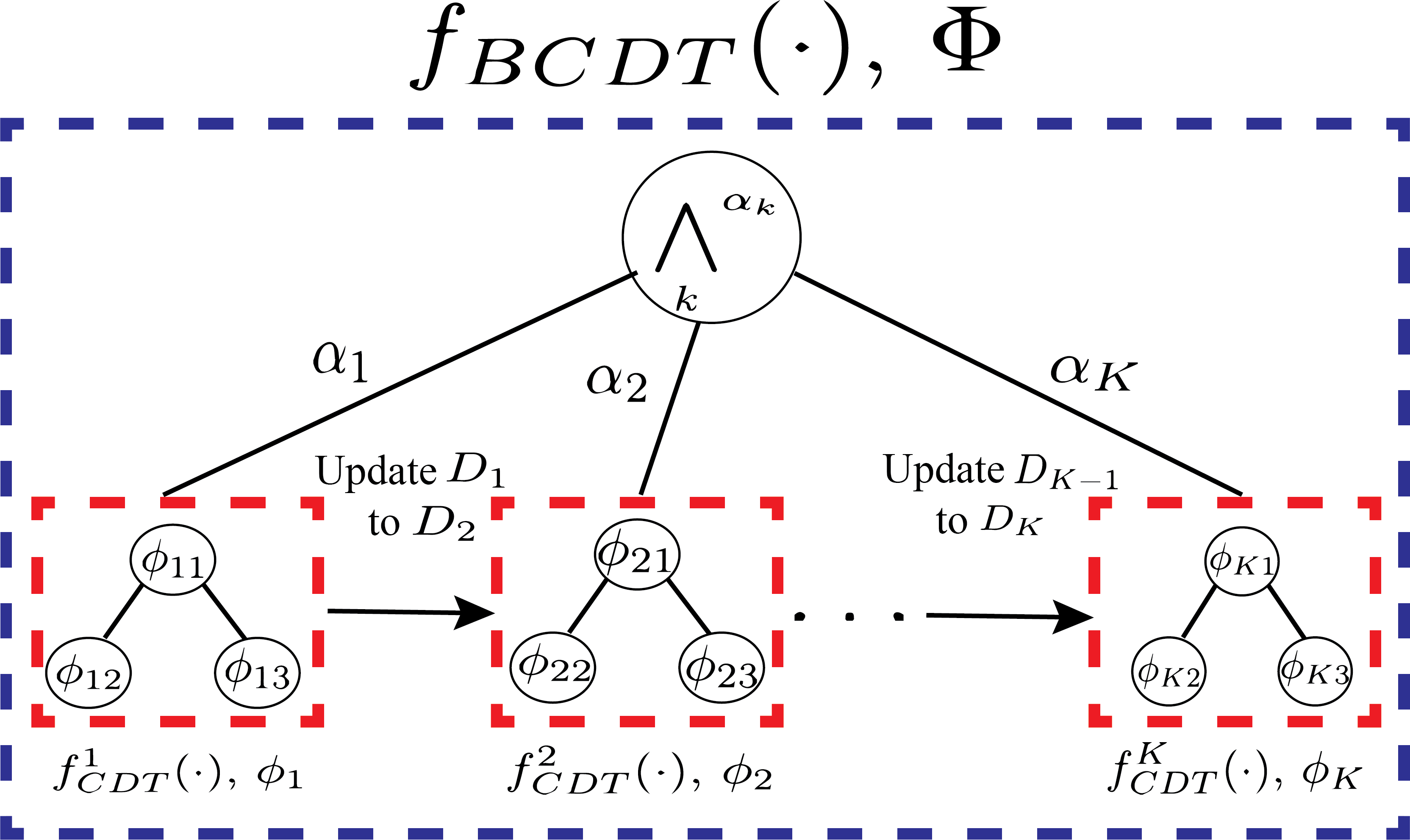}
\caption{Illustration of BCDT Alg.~\ref{alg:boostedtree}. The CDTs and their weights are used in construction of the final classifier $f_{BCDT}(.)$ in Alg.~\ref{alg:boostedtree}, and its corresponding formula $\Phi$. In this figure we have assumed $\forall k \in [0, K]: \alpha_k < M$.}
\label{fig:boosted_example}
\vspace{-4mm}
\end{figure}

The final output $f_{BCDT}(\cdot)$ assigns a label to each data sample.
For simplicity, we abuse notation and consider $C_p = 1$
and $C_n  = -1$, such that $f_{CDT}^{k}(\cdot) \in \{-1, 1\}$ for all $k \in [1, K]$. Note that one of the main assumptions in boosting methods is that each weak learner performs slightly better than random guessing (i.e., coin tossing). Therefore in Alg.~\ref{alg:boostedtree}, if any newly generated tree performs worse than random guessing ($\epsilon_k > 0.5$), we just discard it and generate another tree. An illustration of Alg.~\ref{alg:boostedtree} is shown in Fig.~\ref{fig:boosted_example}.

We use the method from~\cite{bombara2021offline} to convert each CDT $f_{CDT}^k(\cdot)$ to a corresponding STL formula $\phi_k$. 
The output of BCDT method is translated to a set of formulae and associated weights $\{(\phi_k, \alpha_k)\}_{k=1}^{K}$. The STL formula $\Phi = \bigwedge_{k} \phi_k$
is the overall output formula; however, using wSTL \cite{mehdipour2020specifying} we express $\Phi = {\bigwedge_{k}}^{\alpha_k} \phi_k$, to capture the classification performance of each CDT.

\subsection{Construction of Concise Decision Tree}
\label{sec:single_trees}
Decision Trees (DTs)~\cite{breiman1984classification, ripley2007pattern} are sequential decision models with hierarchical structures. In our algorithm, DTs operate on signals with the goal of predicting their labels. Inspired by \cite{bombara2021offline}, we present the Concise Decision Tree (CDT) method $\mathcal{E}$ in Alg.~\ref{alg:dec_tree}, which extends the DT construction algorithm to CDTs, by applying conciseness techniques to generate simpler formulae (detailed in Sec.~\ref{sec:expressivity_methods}).

\begin{algorithm}[htb]
\caption{Concise Decision Tree (CDT) method $\mathcal{E}$}
\begin{algorithmic}[1]
    \State \textbf{Meta-Parameters:} $\mathcal{P}, \mathcal{J}, stop$
    \State \textbf{Input:} $S$, $\phi^{path}$, $h$, $\phi^c_{parent}$
    \State \textbf{Output:} sub-tree $\mathcal{T}$ 
    \State \textbf{if} $stop(\phi^{path}, h, S)$ \textbf{then} \label{alg:line:stop}
    \State \hskip1.5em $c = \mathcal{O} (S, \phi^{path}, \mathcal{P}, h)$ \label{alg:line:leaf_label}
    \State \hskip1.5em \textbf{return} $leaf(c)$ \label{alg:line:leaf}
    \State $\mathcal{T} \leftarrow non\_terminal(\phi^c_{parent})$ \label{alg:line:non_terminal}
    \State $\phi^{new} = \phi^{path} \land \phi^c_{parent}$ \label{alg:line:phi_new}
    \State $S_{\top}, S_{\bot} \leftarrow partition(S, \phi^{new})$ \label{alg:line:partition}
    \State \textbf{for} $\otimes \in \{\top, \bot\}$ \textbf{do} \label{alg:line:for}
    \State \hskip1.5em  $\phi^{c}_\otimes = \mathcal{O}(S_\otimes, \phi^{new}, \mathcal{P}, h+1)$ \label{alg:line:phi_left}
    \State \hskip1.5em $\phi^\otimes = \mathcal{C} (\phi^c_{parent}, \phi^{c}_\otimes, S, \phi^{path}, h)$ \label{alg:line:phi_l}
    \State \hskip1.5em \textbf{if} $\phi^{path} \land \phi^\otimes \succeq^{\mathcal{J}} \phi^{new}$: \label{alg:line:impurity_check}
    \State \hskip3.0em \textbf{return} $\mathcal{E} (S, \phi^{path}, h, \phi^\otimes)$ \label{alg:line:repeat_left}
    \State $\mathcal{T}.left \leftarrow \mathcal{E}(S_{\top}, \phi^{new}, h+1, \phi^c_{\top})$  \label{alg:line:tree_left}
    \State $\mathcal{T}.right \leftarrow \mathcal{E}(S_{\bot}, \phi^{new}, h+1, \phi^c_{\bot})$ \label{alg:line:tree_right}
    \State \textbf{return} $\mathcal{T}$ $\label{alg:line:tree_output}$
\end{algorithmic}
\label{alg:dec_tree}
\end{algorithm}

To limit the complexity of CDTs, we consider three meta-parameters in Alg.~\ref{alg:dec_tree}: (1) PSTL primitives $\mathcal{P}$ capturing the possible ways to split the data at each node, (2) impurity measures $\mathcal{J}$ to select the best primitive at each node, and (3) stop conditions $stop$ to limit the CDTs' growth. The meta-parameters are explained in details in Sec.~\ref{sec:meta_params}.

Alg.~\ref{alg:dec_tree} is recursive, and takes as input (1) the set of labeled signals $S$ at the current node, referred to as \emph{parent} node, (2) the path formula $\phi^{path}$ from the root to the parent node, (3) the depth $h$ from the root to the node, and (4) the candidate formula $\phi^c_{parent}$ for the node. 
At the beginning, the stop conditions $stop$ are checked (line~\ref{alg:line:stop}). If they are satisfied (lines~\ref{alg:line:leaf_label}-\ref{alg:line:leaf}), a single leaf is returned that is marked with label $c$, according to the primitive optimization method in Alg.~\ref{alg:primitive_optimization}. Otherwise, a non-terminal node is created that is associated with the candidate formula $\phi^c_{parent}$ (line~\ref{alg:line:non_terminal}). The formula $\phi^{new}$ is the updated path formula from the root, considering the candidate primitive $\phi^c_{parent}$ of the parent node (line~\ref{alg:line:phi_new}). Next, the data set $S$ is partitioned according to the new formula (line~\ref{alg:line:partition}), where $S_{\top}$ and $S_{\bot}$ are the set of signals that satisfy and violate $\phi^{new}$, respectively.


Following the structure of the tree, first for the left child of the node ($\otimes = \top$) and then for the right child ($\otimes = \bot$), we follow these steps (line~\ref{alg:line:for}): first, the candidate primitive for the child $\phi^{c}_{\otimes}$ is computed from the set $\mathcal{P}$ (line~\ref{alg:line:phi_left}). Then, by applying the conciseness method $\mathcal{C}$ (explained in Sec.~\ref{sec:expressivity_methods}) on the combination of parent's candidate formula $\phi^c_{parent}$ and the child's candidate primitive $\phi^c_{\otimes}$, we find a new formula $\phi^\otimes$ (line~\ref{alg:line:phi_l}) as a new candidate for the parent node. In line~\ref{alg:line:impurity_check}, the notation $\succeq^{\mathcal{J}}$ is used to compare two formulae based on the impurity measure $\mathcal{J}$. If the impurity reduction of the new candidate formula $\phi^\otimes$ is more than the previous candidate $\phi^c_{parent}$, the algorithm is repeated for the parent node, with $\phi^c_{parent}$ replaced by $\phi^\otimes$ (line~\ref{alg:line:repeat_left}). 
Note that the decision tree method in \cite{bombara2021offline} is based on the idea of incremental impurity reduction at each node of the tree. Following the same idea, we argue that by applying the conciseness techniques 
at each node, if the impurity reduction of the new candidate formula is better than the previous one, the new candidate leads to a stronger classifier with a simpler specification. Finally, when there is no more possibility of applying the conciseness method on the parent node, we continue the construction of the tree for the left and right children (lines~\ref{alg:line:tree_left}-\ref{alg:line:tree_right}) and the sub-tree for the parent is returned (line~\ref{alg:line:tree_output}).

\begin{algorithm}[htb]
\caption{Parameterized Primitive Optimization $\mathcal{O}$}
\begin{algorithmic}[1]
    \State \textbf{Meta-Parameters:} $\mathcal{J}, stop$
    \State \textbf{Input:} $S$, $\phi^{path}$, $prim$, $h$
    \State \textbf{Output:} optimal primitive $\phi^*$  
    \State \textbf{if} $stop(\phi^{path}, h, S)$ \textbf{then} \label{opt_alg:line:stop}
    \State \hskip1.5em $\phi^* \leftarrow arg\max_{c \in C} \{p(S, c; \phi^{path})\}$ \label{opt_alg:line:leaf}
    \State \textbf{else} 
    \State \hskip1.5em $\phi^*= \underset{\psi \in prim, \theta \in \Theta}{\argmax} \mathcal{J}(S, partition(S, \phi^{path} \land \psi_\theta))$ 
    \State \textbf{return} $\phi^*$ 
\end{algorithmic}
\label{alg:primitive_optimization}
\end{algorithm}

The parameterized primitive optimization method $\mathcal{O}$, presented in Alg.~\ref{alg:primitive_optimization}, finds the best primitive with optimal evaluation, from the input primitive set $prim$. This method has similar meta parameters as Alg.~\ref{alg:dec_tree} and takes as input (1) the set of labeled signals $S$ at the current node, (2) the path formula $\phi^{path}$ from the root to the current node, (3) a set of input primitives \emph{prim}, and (4) the depth $h$ from the root to the node. If the stop conditions are satisfied (line~\ref{opt_alg:line:stop}), a label $c^* \in C$ is computed (line~\ref{opt_alg:line:leaf}) according the best classification quality, using the partition weight $p(S, c; \phi^{path})$ of the impurity measure (see Sec.~\ref{subsec:impurity});
otherwise, the best primitive from the input primitive set $prim$ is computed by solving an optimization method based on the impurity measure $\mathcal{J}$.

\subsection{Meta Parameters} \label{sec:meta_params}
\subsubsection{PSTL primitives $\mathcal{P}$} 
The splitting rules at each node are simple PSTL formulae, called {\em primitives}~\cite{bombara2021offline}. Here we use {\em first-order primitives} $\mathcal{P}_1$:  $G_{[t_0,t_1]} (s_j \sim \pi)$, $F_{[t_0,t_1]} (s_j \sim \pi)$, where the decision parameters are $(t_0, t_1, \pi)$.

\subsubsection{Impurity measure $\mathcal{J}$} \label{subsec:impurity}

We use the Misclassification Gain (MG) impurity measure \cite{breiman1984classification} as a criterion to select the best primitive at each node. Given a finite set of signals $S$, an STL formula $\phi$, and the subsets of $S$ that are partitioned based on satisfaction of $\phi$ as $S_\top$, $S_\bot=partition (S, \phi)$, we have $MG(S, \{S_\top, S_\bot\}) = MR(S) - \sum_{\otimes \in \{\top, \bot\}} p_\otimes \, MR(S_\otimes)$, where $ MR(S) = \min (p(S, C_p; \phi) \, , \, p(S, C_n; \phi))$, and the $p$ parameters are partition weights computed based on signals' labels and satisfaction of $\phi$.
Here, we extend the robustness-based impurity measures in~\cite{bombara2021offline}
to account for the sample weights $D_k$ from the BCDT in Alg.~\ref{alg:boostedtree}.
The boosted impurity measures are defined by the partition weights below
\begin{equation}
\label{eqn:mg_boosted_weights}
\begin{aligned}
\small
    &{} p_\otimes = \frac{\sum_{(s^i, \ell^i) \in S_\otimes}^{} D_k(s^i) \cdot \rho(\phi, s^i)}{\sum_{(s^i, \ell^i) \in S} D_k(s^i) \cdot |\rho(\phi, s^i)|}, \, \, \, \otimes \in \{\top, \bot\} \\
    &{} p(S, c; \phi) = \frac{\sum_{(s^i, \ell^i) \in S, \, \ell^i = c}^{} D_k(s^i) \cdot |\rho(\phi, s^i)|}{\sum_{(s^i, \ell^i) \in S} D_k (s^i) \cdot |\rho(\phi, s^i)|}
\end{aligned}
\end{equation}
This formulation also works for other types of impurity measures, such as information and Gini gains~\cite{rokach2005}.

\subsubsection{Stop Conditions}
There are multiple stopping conditions that can be considered for terminating Alg.~\ref{alg:dec_tree}.
We stop the growth of trees either when they reach a given depth, or when $\lambda$ percent of the signals belong to the same class. In our implementations, we set $\lambda = 95 \%$.


\subsection{Conciseness} \label{sec:expressivity_methods}
We propose the conciseness method $\mathcal{C}$, presented in Alg.~\ref{alg:concise_algorithm}, to improve the simplicity and interpretability of STL formulae.
This algorithm takes as inputs the candidate primitive $\phi^c_{parent}$ for the parent node, the candidate primitive for its child (either left or right child) $\phi^c_{\otimes}, \otimes \in \{\top, \bot\}$, the set of signals $S$, path formula $\phi^{path}$, and depth $h$ of the parent node. The output of the algorithm is a new candidate primitive for the parent node, denoted by $\phi^c_{new}$. 

First, the method constructs a new PSTL primitive for the parent node, denoted by $\phi_{parent}$, by combining the candidate primitives of the parent and the child nodes (line~\ref{alg:concise:combine}), where the combination operator is denoted by $\biguplus$. This is done by considering the possible ways to combine two candidate primitives, which we propose two heuristic techniques for it. Then, the optimal valuation of the new PSTL primitive is computed by using the optimization method $\mathcal{O}$ and the path formula $\phi^{path}$ (line~\ref{alg:concise:optimization}).

\begin{algorithm}[htb]
\caption{Conciseness Method $\mathcal{C}$}
\begin{algorithmic}[1]
    \State \textbf{Input:} $\phi^c_{parent}, \phi^c_{\otimes}, S$, $\phi^{path}$, $h$
    \State \textbf{Output:} new candidate primitive $\phi^c_{new}$ 
    \State $\phi_{parent} = \phi^c_{parent} \biguplus \phi^c_{\otimes}$ \label{alg:concise:combine}
    \State $\phi^c_{new} = \mathcal{O} (S, \phi^{path}, \phi_{parent}, h)$  \label{alg:concise:optimization}
    \State \textbf{return} $\phi^c_{new}$ 
\end{algorithmic}
\label{alg:concise_algorithm}
\end{algorithm}

The heuristic techniques to combine two primitives and generate shorter PSTL formulae are as following:
\vspace{+1mm}
\subsubsection{Combination of Always operators}
If the candidate primitives of the parent and child nodes are as $\phi^c_{parent} = G_{[t_0, t_1]} \mu_{parent}$ and $\phi^c_{\otimes} = G_{[t_2, t_3]} (\mu_{child})$, respectively, we construct a new PSTL primitive $\phi_{parent} = G_{[t_4, t_5]} ((\mu_{parent}) \land (\mu_{child}))$ for their combination.
For example, given $\phi^c_{parent} = G_{[t_0, t_1]} ((s_1 > \pi_1) \land (s_2 \leq \pi_2))$ and $\phi^c_{\otimes} = G_{[t_2, t_3]} (s_2 > \pi_3)$, the combined PSTL primitive is $\phi_{parent} = G_{[t_4, t_5]} ((s_1 > \pi_1) \land (\pi_3 < s_2 \leq \pi_2))$. 

\subsubsection{Combination of Eventually operators}
Similar to the combination of always operators, if the candidate primitives of the parent and child nodes are as $\phi^c_{parent} = F_{[t_0, t_1]} \mu_{parent}$ and $\phi^c_\otimes = F_{[t_2, t_3]} (\mu_{child})$, respectively, we construct a new PSTL primitive as $\phi_{parent} = F_{[t_4, t_5]} ((\mu_{parent}) \wedge (\mu_{child}))$.

\begin{figure}[h]
\centering
\includegraphics[width=0.95\columnwidth]{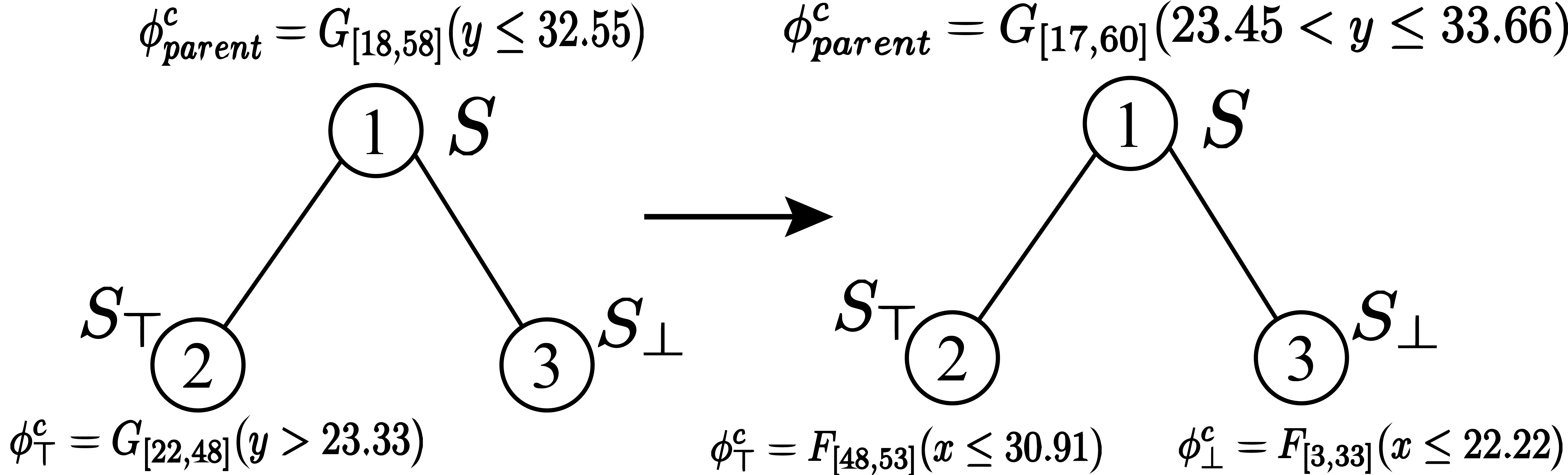}
\caption{Example of applying the conciseness method $\mathcal{C}$ for the naval surveillance data set. On the left, the candidate primitive for the parent node 1 is $\phi^c_{parent} = G_{[18, 58]} (y \leq 32.55)$, and after partitioning the signals, the candidate primitive for the left child (node 2) is computed as $\phi^c_{\top} = G_{[22, 48]} (y > 23.33)$. The conciseness method $\mathcal{C}$ constructs a new candidate PSTL primitive $\phi_{parent} = G_{[t_0, t_1]} (\pi_0 < y \leq \pi_1)$ for their combination, where its optimal valuation according to the primitive optimization method $\mathcal{O}$ is $\phi^\top = G_{[17,60]} (23.45 < y \leq 33.66)$. Due to the higher impurity reduction of $\phi^\top$ compared to $\phi^c_{parent}$, the new candidate $\phi^\top$ is chosen as $\phi^c_{parent}$ for the parent node 1 on the right. The partitioning of the signals and the candidate primitives for the left and right children (nodes 2 and 3) are recomputed according to the new candidate primitive $\phi^c_{parent}$. According to the conciseness technique $\mathcal{C}$, there is no more possibility of combining the candidate primitives of the parent node and its children, because the temporal operators of the parent and its children nodes are different. Therefore, the procedure of CDT construction is followed up from the left and right children.} 
\label{fig:prune_example}
\end{figure}

In Fig.~\ref{fig:prune_example}, we provide an example of how Alg.~\ref{alg:concise_algorithm} works.

\textbf{Remark 1:}
There are multiple ways of combining the primitives to improve interpretability.
For example, given the candidate primitive the parent node as $\phi^c_{parent} = G_{[t_0, t_1]} (\mu_{parent})$, and the candidate primitive of its child as $\phi^c_{\otimes} = F_{[t_2, t_3]} (\mu_{child})$, we can construct a new PSTL primitive for the parent node as $\phi_{parent} = F_{[t_4, t_5]} ((\mu_{parent}) \, U_{[0, t_6]} \, (\mu_{child}))$. We will explore the other ways of combining the primitives in the future works.

\textbf{Remark 2:} Note that our heuristic method combines two primitives whenever their combination improves the impurity measure and classification performance. This leads to a larger set of primitives for constructing the trees, and it is more efficient compared to the naive approach of investigating all possible combinations of primitives.

\subsection{Complexity Analysis}
We denote the lower and the two-sided asymptotic bounds for the complexity of the overall algorithm by $\Omega(.)$ and  $\Theta(\cdot)$, respectively. 
The complexity of the BCDT algorithm (Alg.~\ref{alg:boostedtree}) is equivalent to the complexity of the AdaBoost method with $K$ trees $O(K \, . \, C_{\mathcal{E}}(N))$, where $C_{\mathcal{E}}(N)$ is the complexity of constructing a CDT by Alg.~\ref{alg:dec_tree}, and $N$ is the number of signals to be processed. Let $C_{\mathcal{O}}(N)$ be the complexity of the optimization method in Alg.~\ref{alg:primitive_optimization}. Clearly we have $C_{\mathcal{O}}(N) = \Omega(N)$, because the method must at least check the labels of all signals \cite{cormen2009introduction}. The worst-case complexity of Alg.~\ref{alg:dec_tree} is obtained when at each node the optimal partition has size $(1, N-1)$, and we run the conciseness method (Alg.~\ref{alg:concise:optimization}) for each child of the node, which leads to $2 C_{\mathcal{O}}(N)$. Using the recursive nature of decision trees, the complexity analysis of \cite{bombara2016decision} and the Akra-Bazzi method \cite{cormen2009introduction}, for the worst-case and average-case complexity of $C_{\mathcal{E}}(N)$ we have $\Theta (N + 4 \sum_{k=2}^{N} C_{\mathcal{O}}(k))$ and $\Theta (N \, \cdot \, (1 + 2\int_1^N \frac{C_{\mathcal{O}}(u)}{u^2}) du)$, respectively.

\section{CASE STUDIES}
\label{section:case-studies}
We demonstrate the effectiveness and computational advantages of our method with two case studies. The first is the naval surveillance scenario from Sec.~\ref{sec:motivating-example}. The second is an urban-driving scenario, implemented in the simulator CARLA \cite{dosovitskiy2017carla}. We use Particle Swarm Optimization (PSO) method \cite{kennedy1995particle} for solving the optimization problems in Alg.~\ref{alg:primitive_optimization}. The parameters of the PSO method are tuned empirically. We use $M=100$ in our implementations. We run the case studies on a $3.70$ GHz processor with $16$ GB RAM. 

\subsection{Naval Surveillance}
\label{sec:naval_scenario}
\begin{figure}[h]
\centering
    \subfigure[]
    {\includegraphics[width=0.49\columnwidth]{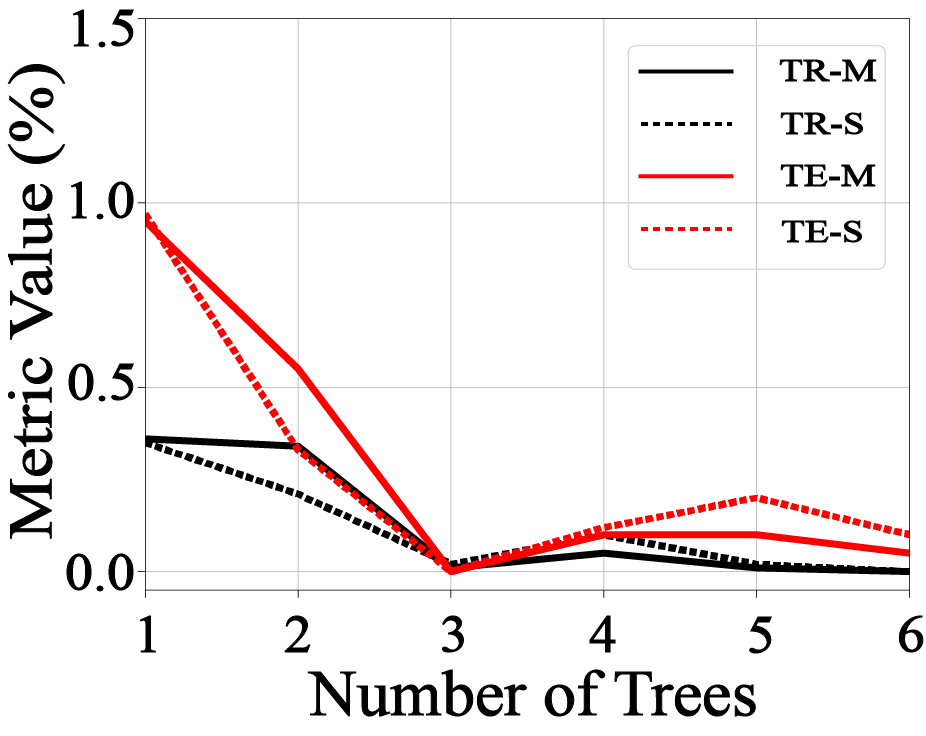}\label{fig:naval_tr_te}}
    \subfigure[]
    {\includegraphics[width=0.49\columnwidth]{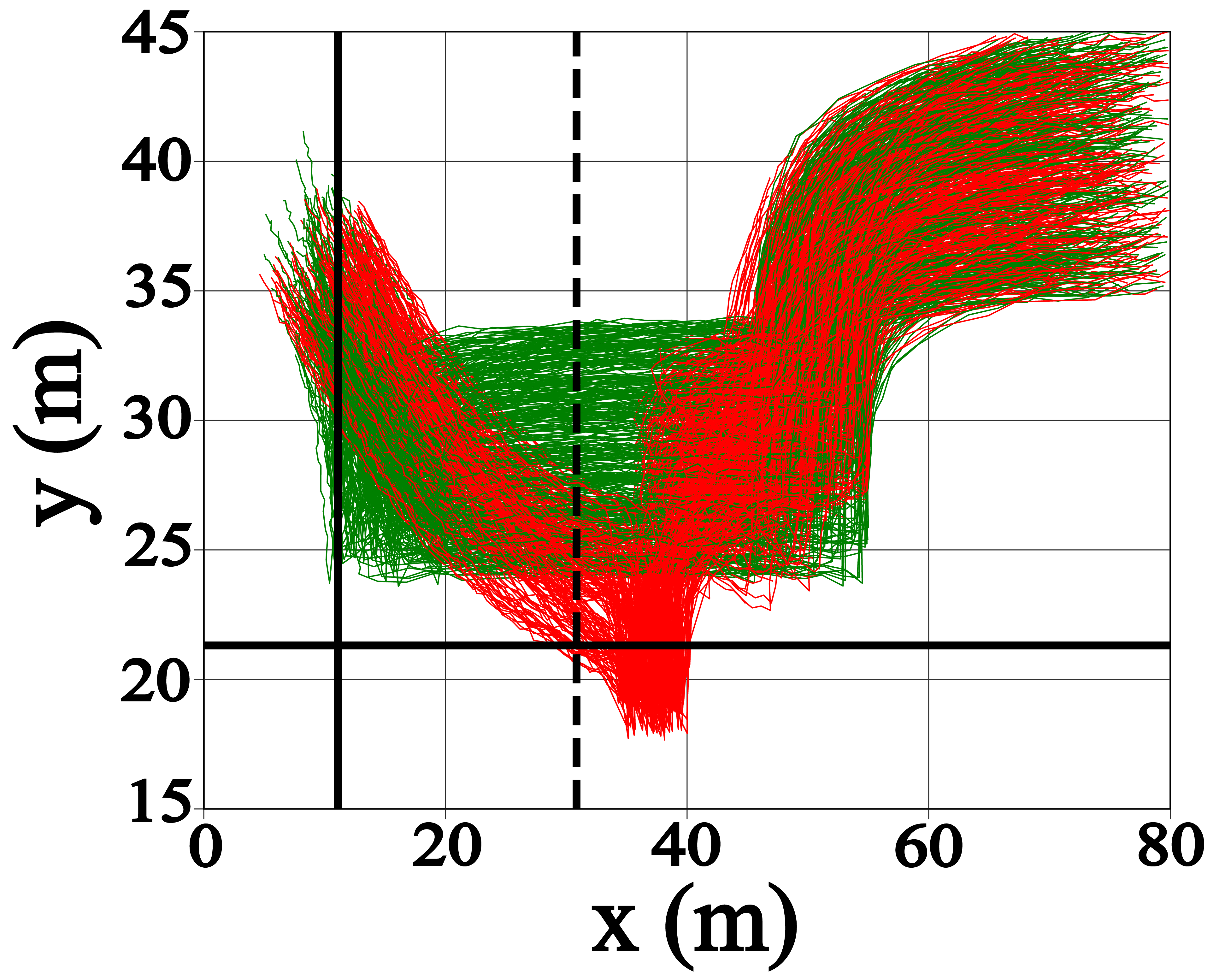}\label{fig:naval_traj}}
    \caption{(a) Classification performance of our framework for naval surveillance scenario, with different number of decision trees $K$. The best performance is obtained with $K = 3$, (b) Examples of trajectories from the naval surveillance case study. The green and red trajectories belong to normal and anomalous behaviors, respectively. For formula $\Phi^{Naval}_1$, the thresholds of the always and eventually operators are shown by solid and dashed black lines, respectively.}
\vspace{-2mm}
\end{figure}

We compare our inference algorithm with the methods from \cite{bombara2021offline} (the DTL4STL tool) and \cite{mohammadinejad2020interpretable}. The dataset is composed of 2000 signals, with 1000 normal and 1000 anomalous trajectories. Each signal has 61 timepoints (see Fig.~\ref{fig:naval_traj} for some example trajectories). We test our algorithm with 5-fold cross validation and maximum depth = 3 for the trees (as in \cite{bombara2021offline}). The results are provided in Table.~\ref{table:naval_results} for different number of decision trees $K$ in Alg.~\ref{alg:boostedtree}; TR-M$(\%)$ and TR-S$(\%)$ are the mean and standard deviation of the MCR in the training phase, respectively; TE-M$(\%)$ and TE-S$(\%)$ are the mean and standard deviation of the MCR in the test phase; R is the runtime, and CT is the number of times that by applying the conciseness method $\mathcal{C}$ during the construction of CDTs, a simpler formula is found.

In Fig.~\ref{fig:naval_tr_te}, the classification performance of our framework is represented, with respect to different number of decision trees $K$. From this figure and Table.~\ref{table:naval_results} it is clear that the best classification performance, over both training and test phases, is obtained with $K = 3$, where we find a set of concise trees that are able to classify all signals correctly in the test phase. Note that adding to the number of trees increases the complexity of the framework and leads to capturing finer details of the dataset, which has the risk of overfitting, as the TE-M increases for $K > 3$ (see Fig.~\ref{fig:naval_tr_te}).

\begin{table}[htb]
\scriptsize
\caption{}
\vspace{-4mm}
\label{table:naval_results}
\begin{center}
\begin{tabular}{|c|c|c|c|c|c|c|}
\hline
$K$ & TR-M (\%) & TR-S (\%) & TE-M (\%) & TE-S (\%) & R & CT \\
\hline
1 & 0.36 & 0.35 & 0.95 & 0.97 & 11m 8s & 4 \\
\hline
2 & 0.34 & 0.21 & 0.55 & 0.33 & 30m 47s & 14 \\
\hline
3 & 0.01 & 0.02 & 0.00 & 0.00 & 33m 16s & 10 \\ 
\hline
4 & 0.05 & 0.10 & 0.10 & 0.12 & 61m 33s & 29\\
\hline
5 & 0.01 & 0.02 & 0.10 & 0.20 & 81m 52s & 33\\
\hline
6 & 0.00 & 0.00 & 0.05 & 0.10 & 85m 55s & 38\\
\hline
\end{tabular}
\end{center}
\vspace{-4mm}
\end{table}

In the following, the learned formulae with $K = 3$ are presented over all the folds. At each fold $f$, first the initial wSTL formula $\Phi_f$ learned by Alg.~\ref{alg:boostedtree} is presented, which is the weighted conjunction of three STL formulae; then by applying the heuristic technique from Alg.~\ref{alg:boostedtree} for trees with weight $M$, the final output $\Phi^{Naval}_f$ is presented:

\begin{itemize}
    \item Fold 1: $\Phi_1 = \phi_{11}^M \wedge \phi_{12}^{2.71} \wedge \phi_{13}^{2.88}$
    \vspace{+1mm}
    
    $\quad \quad \Rightarrow \Phi^{Naval}_1 = \phi_{11} = \phi_{111} \wedge \phi_{112}$
    \vspace{+1mm}
    
    $\quad \quad \phi_{111} = F_{[28, 53]} (x \leq 30.85)$
    \vspace{+1mm}
    
    $\quad \quad \phi_{112} =  G_{[2, 26]} ((y > 21.31) \wedge (x > 11.10))$
    \vspace{+2mm}
    
    \item Fold 2: $\Phi_2 = \phi_{21}^{2.59} \wedge \phi_{22}^{0.80} \wedge \phi_{23}^M$
    \vspace{+1mm}
    
    $\quad \quad \Rightarrow \Phi^{Naval}_2 = \phi_{23} = \phi_{231} \wedge \phi_{232}$
    \vspace{+1mm}
    
    {\small $\quad \phi_{231} = G_{[8, 32]} (y > 23.15), \, \phi_{232} =  F_{[20, 55]} (x \leq 33.57)$}
    \vspace{+2mm}
    
    \item Fold 3: $\Phi_3 = \phi_{31}^M \wedge \phi_{32}^{2.49} \wedge \phi_{33}^{2.64}$
    \vspace{+1mm}
    
    $\quad \quad \Rightarrow \Phi^{Naval}_3 = \phi_{31} = \phi_{311} \wedge \phi_{312}$
    \vspace{+1mm}
    
    {\small $\quad \phi_{311} = F_{[57, 60]} (x \leq 33.91), \, \phi_{312} =  G_{[7, 51]} (y > 21.36)$}
    \vspace{+2mm}
    
    \item Fold 4: $\Phi_4 = \phi_{41}^{2.65} \wedge \phi_{42}^M \wedge \phi_{43}^M$
    \vspace{+1mm}
    
    $\quad \quad \Rightarrow \Phi^{Naval}_4 = \phi_{43} = \phi_{431} \wedge \phi_{432}$
    \vspace{+1mm}
    
    $\quad \quad \phi_{431} = F_{[52, 55]} (x \leq 35.34)$
    \vspace{+1mm}
    
    $\quad \quad \phi_{432} =  G_{[0, 26]} ((y > 22.20) \wedge (x > 11.73))$
    \vspace{+2mm}
    
    \item Fold 5: $\Phi_5 = \phi_{51}^{2.53} \wedge \phi_{52}^M \wedge \phi_{53}^{1.83}$
    \vspace{+1mm}
    
    $\quad \quad \Rightarrow \Phi^{Naval}_5 = \phi_{52} = \phi_{521} \wedge \phi_{522}$
    \vspace{+1mm}
    
    {\small $\quad \phi_{521} = G_{[9, 44]} (y > 23.43), \, \phi_{522} =  F_{[57, 59]} (x \leq 33.10)$}
\end{itemize}


Note that there are some similarities between the inferred formulae in different folds; for example, the structures of the formulas $\phi_{112}$ and $\phi_{432}$ are the same and their thresholds and time bounds are really close to each other.
Also it is worth to mention that in the fourth fold, although both formulae $\phi_{42}$ and $\phi_{43}$ have weight $M$, formula $\phi_{43}$ is chosen over $\phi_{42}$ because $\phi_{43}$ has less number of Boolean and temporal operators (see Sec.~\ref{sec:boosted_trees}). The output formulae of each fold are simple and easy to interpret. For example, from the plain English translation of formula $\Phi^{Naval}_1$, the behavior of normal vessels is interpreted as: "Normal vessel's $x$ and $y$ coordinates are bigger than 11.10 and 21.31m, respectively, over the time interval $[2, 26]$, and their x coordinate gets less than or equal to 30.85m, at some timepoint in the time interval [28, 53]". The thresholds of the formula $\Phi^{Naval}_1$ are shown in Fig.~\ref{fig:naval_traj}.

In \cite{bombara2021offline}, using first-order primitives and maximum tree depth of 3, the authors get a MCR with mean 1.3$\%$ and standard deviation 0.28$\%$ for this data set. To provide a fair comparison, we ran the algorithm from \cite{bombara2021offline} on the same
computer that we used for our algorithm and for the same data set. 
We obtained a MCR with mean $1.5 \%$ and standard deviation $0.5 \%$ in the test phase, with total runtime of 33 seconds. An example formula learned in one of the folds using the method 
from \cite{bombara2021offline} is:
\begin{align} \nonumber
\small
    &{}(\phi_1 \wedge \phi_2) \vee (\neg \phi_1 \wedge ((\phi_3 \wedge \phi_4) \vee (\neg \phi_3 \wedge \phi_5))) \\ \nonumber 
    &{}\phi_1 = G_{[39, 60]} (x \leq 19.5), \quad \phi_2 = F_{[11, 38]} (x > 41.2) \\ \nonumber 
    &{}\phi_3 = G_{[20, 59]} (y < 32.3), \quad \phi_4 = G_{[59, 60]} (x \leq 39.2) \\ \nonumber 
    &{}\phi_5 = G_{[20, 53]} (y > 29.7)
\end{align}

Compared to \cite{bombara2021offline}, our algorithm obtains a better classification performance, in addition to simpler and more interpretable formulae, at the cost of higher runtime due to the boosting and conciseness techniques. In \cite{mohammadinejad2020interpretable}, the authors obtain a MCR with mean $5\%$ in test phase and total runtime of 45 minutes and the formula learned in their work is $(y \geq 19.74) U_{[0, 9.84]} (x \leq 24.86)$. From the interpretability view, both the formulae learned by our algorithm and by \cite{mohammadinejad2020interpretable} are simple and easy to interpret and both methods have roughly similar runtime, but our algorithm has noticeably better classification performance. 

\subsection{Urban Driving}
\label{sec:urban_scenario}
\begin{figure}[h]
\centering
\includegraphics[width=0.8\columnwidth]{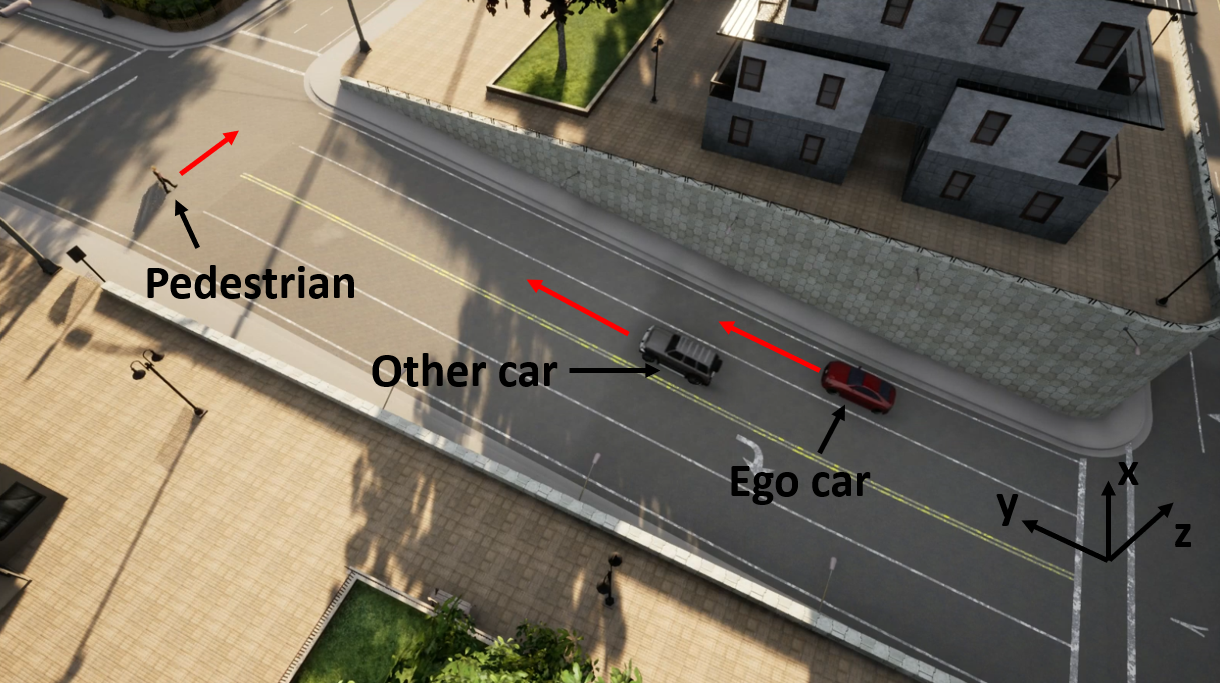}
\caption{Urban driving scenario implemented in the simulator CARLA \cite{dosovitskiy2017carla}} 
\label{fig:carla-scenario}
\vspace{-2mm}
\end{figure}

Consider an autonomous vehicle (referred to as \emph{ego}) driving in an urban environment shown in Fig.~\ref{fig:carla-scenario}. The scenario also contains a pedestrian and another car, which is assumed to be driven by a "reasonable" human who obeys traffic laws. Ego and the other car are in different, adjacent lanes, moving in the same direction. The cars move uphill in the $y-z$ plane of the coordinate frame, towards positive $y$ and $z$ directions,
with no lateral movement in the $x$ direction. The accelerations of the cars are constant, and smaller for ego.

The positions and accelerations of the cars are initialized such the other car is always ahead of ego. The vehicles are headed towards an intersection without any traffic lights. There is an unmarked cross-walk at the end of the road before the intersection. When the pedestrian crosses the street, the other car brakes to stop before the intersection. If the pedestrian does not cross, the other car keeps moving without decreasing its velocity. 
Ego does not have a clear line-of-sight to the pedestrian crossing at the intersection, because of the other car and the uphill shape of the road. The goal is to develop a method allowing ego to infer whether a pedestrian is crossing the street by observing the behavior (e.g., relative position and velocity over time) of the other car.

\begin{figure} [h] 
    \centering
    \subfigure[]
    {\includegraphics[width=0.51\columnwidth]{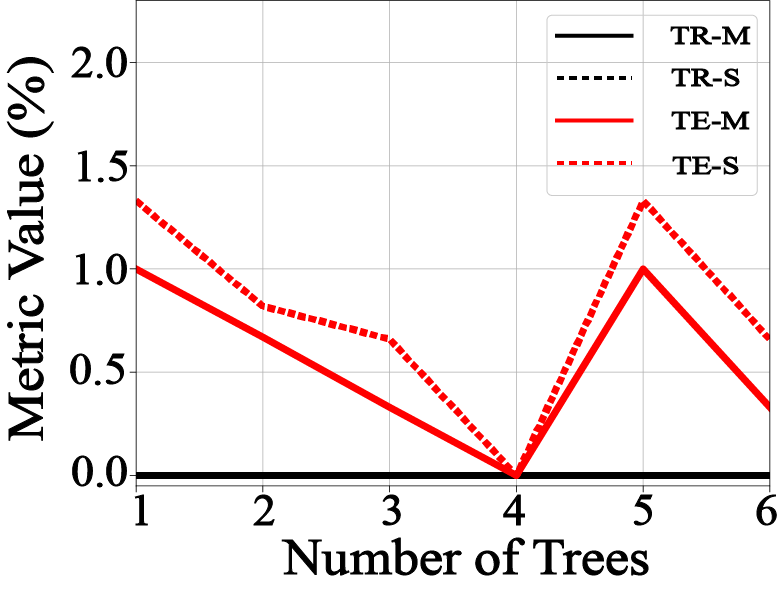}\label{fig:carla-tr_te}}
    \subfigure[]
    {\includegraphics[width=0.49\columnwidth, height=0.40\columnwidth]{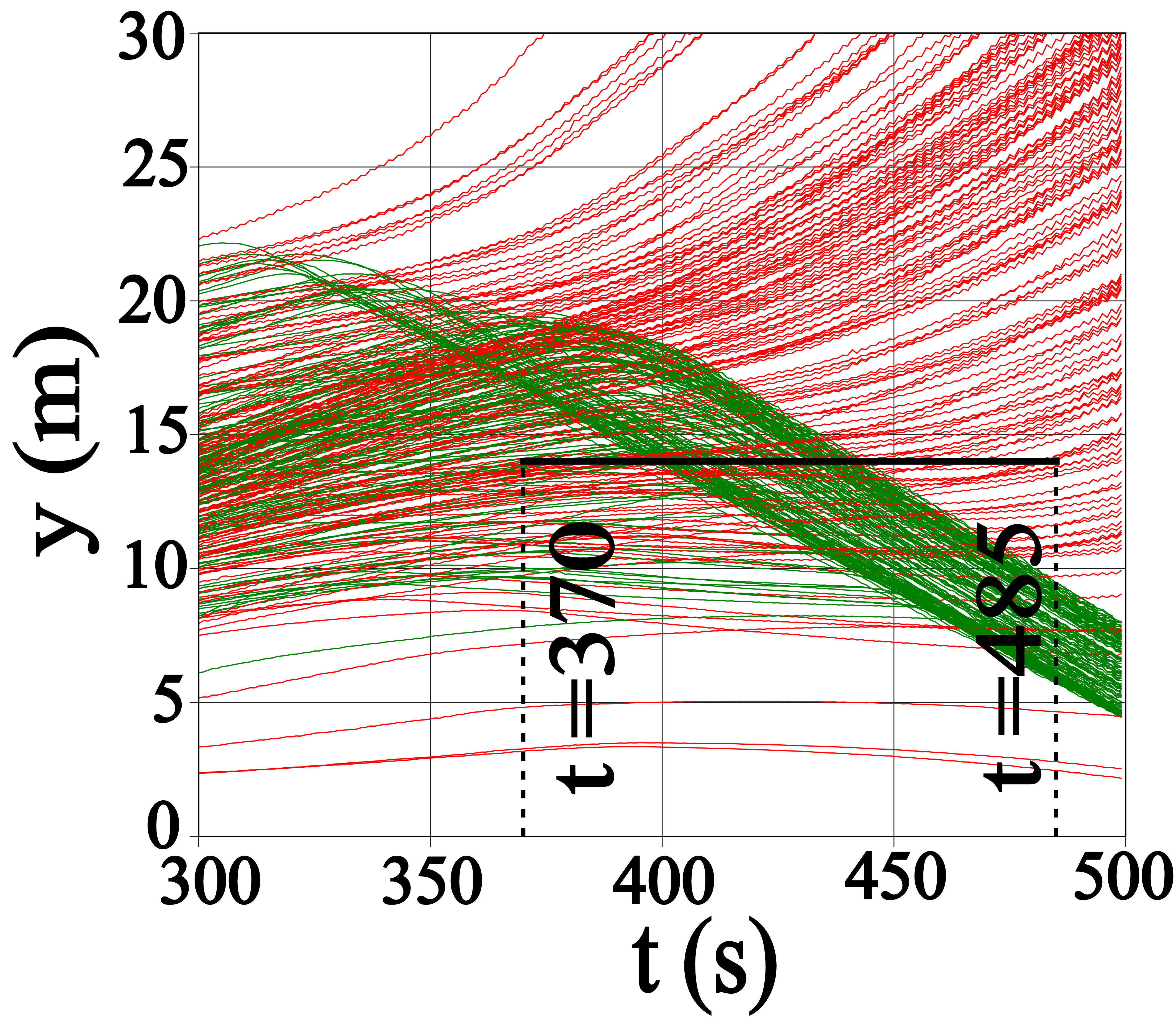}\label{fig:carla-y}}
    \subfigure[]
    {\includegraphics[width=0.49\columnwidth, height=0.40\columnwidth]{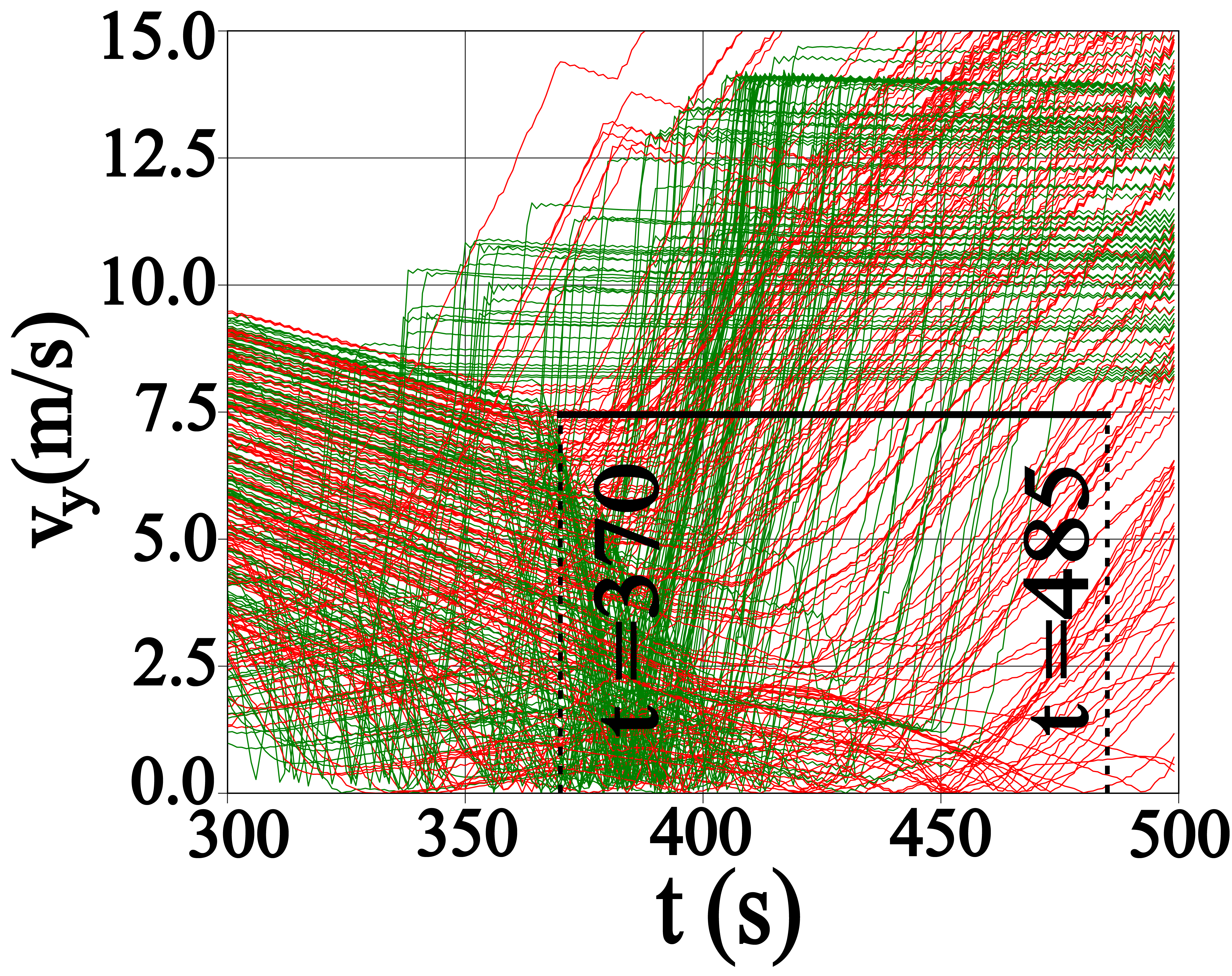}\label{fig:carla-v-y}} 
    \caption{(a) Our framework's classification performance for urban driving scenario, with different number of decision trees $K$. The best performance is obtained with $K = 4$, (b) and (c) the y component of relative distance and relative velocity between ego and the other vehicle, respectively, over time. The green and red signals belong to the cases when there is a pedestrian crossing the street and when there is no pedestrian crossing, respectively. The thresholds and time bound of the formula $\Phi^{Urban}_3$ are shown by solid black lines.}
\vspace{-2mm}
\end{figure}

The simulation of this scenario ends whenever ego gets closer than 8$m$ to the intersection. We assume that labeled behaviors (relative distances and velocities)
are available, where the labels indicate whether a pedestrian is crossing or not. We collected 300 signals with 500 uniform time-samples per trace, where 150 were with and 150 without pedestrians crossing the street (see Fig.~\ref{fig:carla-y} and~\ref{fig:carla-v-y}). The dataset is available in \cite{aasi2022carla}.
We evaluate our algorithm with 5-fold cross-validation and maximum depth = 2 for the trees. The results are shown in Table~\ref{table:carla_results} for different values of $K$. 

\begin{table}[htb]
\scriptsize
\caption{}
\vspace{-4mm}
\label{table:carla_results}
\begin{center}
\begin{tabular}{|c|c|c|c|c|c|c|}
\hline
$K$ & TR-M (\%) & TR-S (\%) & TE-M (\%) & TE-S (\%) & R & CT \\
\hline
1 & 0.00 & 0.00 & 1.00 & 1.33 & 7m 10s & 2 \\
\hline
2 & 0.00 & 0.00 & 0.67 & 0.82 & 9m 57s & 2 \\
\hline
3 & 0.00 & 0.00 & 0.33 & 0.66 & 14m 52s & 1 \\ 
\hline
4 & 0.00 & 0.00 & 0.00 & 0.00 & 24m 40s & 3\\
\hline
5 & 0.00 & 0.00 & 1.00 & 1.33 & 24m 49s & 3\\
\hline
6 & 0.00 & 0.00 & 0.33 & 0.66 & 32m 52s & 3\\
\hline
\end{tabular}
\end{center}
\vspace{-4mm}
\end{table}

The classification performance of our framework is shown in Fig.~\ref{fig:carla-tr_te}, for different number of decision trees $K$, and the best performance is obtained with $K = 4$. In the following the learned formulae with $K = 4$ are presented, by following the same notation as naval surveillance scenario in Sec.~\ref{sec:naval_scenario}:

\begin{itemize}
    \item Fold 1: $\Phi_1 = \phi_{11}^M \wedge \phi_{12}^{2.74} \wedge \phi_{13}^{M} \wedge \phi_{14}^M$
    \vspace{+1mm}
    
    $\quad \quad \Rightarrow \Phi^{Urban}_1 = \phi_{13} = \phi_{131} \wedge \phi_{132}$
    \vspace{+1mm}
    
    {\small $\phi_{131} = F_{[463, 499]} (y \leq 8.78) , \, \phi_{132} =  G_{[477, 481]} (v_y > 8.01)$}
    \vspace{+2mm}
    
    \item Fold 2: $\Phi_2 = \phi_{21}^{2.65} \wedge \phi_{22}^{M} \wedge \phi_{23}^M \wedge \phi_{24}^M$
    \vspace{+1mm}
    
    $\quad \quad \Rightarrow \Phi^{Urban}_2 = \phi_{23} = \phi_{231} \wedge \phi_{232}$
    \vspace{+1mm}
    
    {\small $\phi_{231} = F_{[463, 488]} (z \leq 1.40), \, \phi_{232} =  G_{[488, 493]} (v_z > 1.19)$}
    \vspace{+2mm}
    
    \item Fold 3: $\Phi_3 = \phi_{31}^M \wedge \phi_{32}^{M} \wedge \phi_{33}^{M} \wedge \phi_{34}^M$
    \vspace{+1mm}
    
    $\quad \quad \Rightarrow \Phi^{Urban}_3 = \phi_{32}$
    \vspace{+1mm}
    
    $\phi_{32} = F_{[370, 485]} ((y \leq 14.01) \wedge (v_y > 7.45))$
    \vspace{+2mm}
    
    \item Fold 4: $\Phi_4 = \phi_{41}^{3.03} \wedge \phi_{42}^M \wedge \phi_{43}^M \wedge \phi_{44}^M$
    \vspace{+1mm}
    
    $\quad \quad \Rightarrow \Phi^{Urban}_4 = \phi_{44} = \phi_{441} \wedge \phi_{442}$
    \vspace{+1mm}
    
    $\quad \quad \phi_{441} = F_{[474, 486]} (y \leq 9.47)$
    \vspace{+1mm}
    
    $\quad \quad \phi_{442} =  G_{[476, 496]} ((v_z > 1.03) \wedge (y > 2.65))$
    \vspace{+2mm}
    
    \item Fold 5: $\Phi_5 = \phi_{51}^{M} \wedge \phi_{52}^M \wedge \phi_{53}^{2.47} \wedge \phi_{54}^M$
    \vspace{+1mm}
    
    $\quad \quad \Rightarrow \Phi^{Urban}_5 = \phi_{51}$
    \vspace{+1mm}
    
    $\quad \phi_{51} = F_{[384, 469]} ((y \leq 49.80) \wedge (v_y > 9.13))$
\end{itemize}
Notice that the main objective of this scenario is to infer whether a pedestrian is crossing the street, based on the behavior of the other car when it gets close to the intersection. Hence, we expect the desired specifications to be short and they reason over the signals at time intervals close to the end of the simulation. This conforms to the time intervals of our inferred formulae and the fact that at each fold, we have trees with perfect classification in training phase (weight $M$). The output formulae of our method are simple and easy to understand. For example, $\Phi^{Urban}_3$ states that there is a pedestrian crossing the street, if "at some timepoint in the time interval [370, 485], the vehicles get closer than 14.01m in the $y-$direction, and the $y$ component of ego's velocity gets bigger than the corresponding component of other car by 7.45m/s". This simply means that the other car is stopped at the intersection, because a pedestrian is crossing it, and ego is getting close to the other car; therefore, in the $y-$direction, the relative distance gets smaller and the velocity of ego gets bigger than the other car. The thresholds of formula $\Phi^{Urban}_3$ are shown in Fig.~\ref{fig:carla-y} and~\ref{fig:carla-v-y}.


To provide a fair comparison, we evaluate the performance of the algorithm from \cite{bombara2021offline} on the same data set and on the same computer that is used for the algorithm developed in this paper. For the algorithm in \cite{bombara2021offline}, with first-order primitives, 5-fold cross validation and maximum depth of 2 for the trees, we obtained a mean MCR of 1$\%$ with standard deviation 1.5$\%$ in the test phase, with total runtime of 7.72 seconds. An example formula learned in one of the folds using the method from \cite{bombara2021offline} is $F_{[474, 499]} (z < 1.2) \wedge F_{[0, 499]} (v_y > 8.97)$. The results show that our inferred formulae either have the same structure or are simpler than the formulae inferred by \cite{bombara2021offline}. Moreover, our method achieves better classification performance than the algorithm in \cite{bombara2021offline}, at the cost of higher execution time. 

\section{CONCLUSION} \label{section:conclusion}
In this paper, we propose a novel method for two-class classification of time-series data. Our algorithm grows an ensemble of decision trees that are empowered by conciseness techniques, to improve the interpretability of the formulae. The classification and interpretability advantages of our algorithm are evaluated on naval surveillance and urban-driving case studies, and are compared with two algorithms from literature. In future works, we will investigate alternate ways of achieving a tradeoff between formula conciseness and MCR performance, with faster execution time. Moreover, we will consider the STL inference from signals with heterogeneous time lengths.

\bibliographystyle{IEEEtran}
\bibliography{references}

\end{document}